\documentclass{article}

\usepackage{microtype}
\usepackage{graphicx}
\usepackage{subfigure}
\usepackage{booktabs} 

\usepackage{multicol}

\usepackage{hyperref}
\usepackage{mdframed}
\graphicspath{{images/}}

\usepackage{amsmath,amssymb,amsthm}
\usepackage[nonatbib,preprint]{neurips_2019}

\usepackage[utf8]{inputenc} % allow utf-8 input
\usepackage[T1]{fontenc}    % use 8-bit T1 fonts
\usepackage{hyperref}       % hyperlinks
\usepackage{url}            % simple URL typesetting
\usepackage{booktabs}       % professional-quality tables
\usepackage{amsfonts}       % blackboard math symbols
\usepackage{nicefrac}       % compact symbols for 1/2, etc.
\usepackage{microtype}      % microtypography
\usepackage{graphicx}
\usepackage{amsmath}

\title{End-to-End Learning of Geometric Deformations of Feature Maps for Virtual Try-On}

\author{
  Thibaut Issenhuth, Jérémie Mary, Clément Calauzènes\\
  Criteo AI Lab\\
  Paris, France \\
  \texttt{ \{t.issenhuth, j.mary, c.calauzenes\}@criteo.com} 
}

\usepackage{expl3}
\ExplSyntaxOn
\newcommand\latinabbrev[1]{
  \peek_meaning:NTF . {% Same as \@ifnextchar
    #1\@}%
  { \peek_catcode:NTF a {% Check whether next char has same catcode as \'a, i.e., is a letter
      #1.\@ }%
    {#1.\@}}}
\ExplSyntaxOff

\def\ie{\latinabbrev{i.e}}

\begin{document}

\maketitle

\begin{abstract}
 The 2D virtual try-on task has recently attracted a lot of interest from the research community, for its direct potential applications in online shopping as well as for its inherent and non-addressed scientific challenges. This task requires to fit an in-shop cloth image on the image of a person. It is highly challenging because it requires to warp the cloth on the target person while preserving its patterns and characteristics, and to compose the item with the person in a realistic manner. Current state-of-the-art models generate images with visible artifacts, due either to a pixel-level composition step or to the geometric transformation. In this paper, we propose WUTON: a Warping U-net for a Virtual Try-On system. It is a siamese U-net generator whose skip connections are geometrically transformed by a convolutional geometric matcher. The whole architecture is trained end-to-end with a multi-task loss including an adversarial one. This enables our network to generate and use realistic spatial transformations of the cloth to synthesize images of high visual quality. The proposed architecture can be trained end-to-end and allows us to advance towards a detail-preserving and photo-realistic 2D virtual try-on system. Our method  outperforms the current state-of-the-art with visual results as well as with the Learned Perceptual Image Similarity (LPIPS) metric.\footnote{Code and implementation details are available at \textsc{anonymized} and supplementary material } 
\end{abstract}
\section{Introduction}
A photo-realistic virtual try-on system would be a significant improvement for online shopping. Whether used to create catalogs of new products or to propose an immersive environment for shoppers, it could impact e-shop and open the door for new easy image editing possibilities. The training data we consider is made of \emph{paired} images that are made of the picture of one cloth and the same cloth worn by a model. Then providing an \emph{unpaired} tuple of images: one picture of cloth and one picture of a model with a different cloth, we aim to replace the cloth worn by the model.

An early line of work  addressed this challenge using  3D measurements and model-based methods \cite{guan2012drape,hahn2014subspace,pons2017clothcap}. However, these are by nature computationally intensive and require expensive material, which would not be acceptable at scale for shoppers. Recent works aim to leverage deep generative models to tackle the virtual try-on problem \cite{jetchev2017conditional,han2018viton,wang2018toward,dong2019towards}. CAGAN \cite{jetchev2017conditional} proposes a U-net based Cycle-GAN \cite{isola2017image} approach. However, this method fails to generate realistic results since these networks cannot handle large spatial deformations. In VITON \cite{han2018viton}, the authors use the shape context matching algorithm \cite{belongie2002shape} to warp the cloth on a target person and learn an image composition with a U-net generator. To improve this model, CP-VTON \cite{wang2018toward} incorporates a convolutional geometric matcher \cite{rocco2017convolutional} which learns the parameters of geometric deformations (i.e thin-plate spline transform  \cite{bookstein1989principal}) to align the cloth with the target person. In MG-VTON \cite{dong2019towards}, the task is extended to a multi-pose try-on system, which requires to modify the pose as well as the upper-body cloth of the person.

In this second line of approach, a common practice is to use what we call a \emph{human parser} which is a pre-trained system able to segment the area to replace on the model pictures: the upper-body cloth as well as neck and arms. In the rest of this work, we also assume this parser to be known.

The recent methods for a virtual try-on struggle to generate realistic spatial deformations, which is necessary to warp and render clothes with complex patterns. Indeed, with solid color tees, unrealistic deformations are not an issue because they are not visible. However, for a tee-shirt with stripes or patterns, it will produce unappealing images with curves, compressions and decompressions. Figure \ref{fig:comparison} shows these kinds of unrealistic geometric deformations generated by CP-VTON \cite{wang2018toward}.

To alleviate this issue, we propose an end-to-end model composed of two modules: a convolutional geometric matcher \cite{rocco2017convolutional} and a siamese U-net generator. 
We train end-to-end, so the geometric matcher benefits from the losses induced by the final synthetic picture generation. 
Our architecture removes the need for a final image composition step and generates images of high visual quality with realistic geometric deformations. Main contributions of this work are:  
\begin{itemize}
    \item We propose a simple end-to-end architecture able to generate realistic deformations to preserve complex patterns on clothes such as stripes. This is made by back-propagating the loss computed on the final synthesized images to a learnable geometric matcher. 
    \item We suppress the need for a final composition step present in the best current approaches such as \cite{wang2018toward} using an adversarially trained generator. This performs better on the borders of the replaced object and provides a more natural look of shadows and contrasts.
    \item We show that our approach significantly outperforms the state-of-the-art with visual results and with a quantitative metric measuring image similarity, LPIPS \cite{zhang2018unreasonable}.
    \item We identify the contribution of each part of our net by an ablation study. Moreover, we exhibit a good resistance to low-quality human parser at inference time. 
\end{itemize}
\section{Problem statement and related work}
Given the 2D images $p \in \mathbb{R}^{h\times w \times 3}$ of a person and $c \in \mathbb{R}^{h\times w \times3}$ of a clothing item, we want to generate the image $\Tilde{p} \in \mathbb{R}^{h\times w\times 3}$ where the person from $p$ wears the cloth from $c$. The task can be separated in two parts : the geometric deformation $T$ required to align $c$ with $p$, and the refinement that fits the aligned cloth $\Tilde{c} = T(c)$ on $p$. These two sub-tasks can be modelled with learnable neural networks, i.e spatial transformers networks $STN$ \cite{jaderberg2015spatial, rocco2017convolutional} that output parameters $\theta = STN(p,c)$ of geometric deformations, and conditional generative networks $G$ that give $\Tilde{p} = G(p,c,\theta)$. 

Because it would be costly to construct a dataset with $(p,c,\Tilde{p})$ triplets, previous works \cite{han2018viton,wang2018toward} propose to use an agnostic person representation $ap \in \mathbb{R}^{h\times w\times c}$ where the clothing items in $p$ are hidden but identity and shape of the persons are preserved. $ap$ is built with pre-trained human parsers and pose estimators : $ap = h(p) $. These triplets $(ap,c,p)$ allow to train for reconstrution. $(ap,c)$ are the inputs, $\Tilde{p}$ the output and $p$ the ground-truth. We finally have the conditional generative process :
\begin{equation}
    \Tilde{p} = G(\underbrace{h(p)}_{\text{agnostic person}},\underbrace{c}_{\text{cloth}},\underbrace{STN(h(p),c)}_{\text{geometric transform}})
\end{equation}
\paragraph{\textbf{Conditional image generation.}} Generative models for image synthesis have shown impressive results with the arrival of adversarial training \cite{goodfellow2014generative}.
Combined with deep networks \cite{radford2015unsupervised}, this approach has been extended to conditional image generation in \cite{mirza2014conditional} and performs increasingly well on a wide range of tasks, from image-to-image translation \cite{isola2017image,zhu2017unpaired} to video editing \cite{shetty2018adversarial}. However, these models cannot handle large spatial deformations and fail to modify the shape of objects \cite{mejjati2018unsupervised}, which is necessary for a virtual try-on system. %, handling these spatial deformations is necessary  : the cloth has to be aligned with the body shape of the person.
\paragraph{\textbf{Image composition.}} Recent approaches for image composition combine STNs \cite{jaderberg2015spatial} with GANs to align and merge two images. In \cite{lin2018st}, Lin et al. use a sequence of warp generated by an STN to place a foreground object in a background image. 
Recently, SF-GAN \cite{zhan2018spatial} separated the task in two stages: an STN warping the object, and a refinement network adapting the texture and appearance of the object.

\paragraph{\textbf{Geometric deformations in generative models.}} The problem of handling large spatial deformations in generative models has mainly been studied in the context of pose-guided person image generation. This task consists in generating a person image, given a reference person and a target pose. Some approaches use disentanglement to separate pose and shape information from appearance, which allows reconstructing a reference person in a different pose \cite{ma2018disentangled,lorenz2019unsupervised,esser2018variational}. 
However, recent state-of-the-art approaches for pose-guided person generation include explicit spatial transformations in their architecture, whether learnt \cite{jaderberg2015spatial} or not. In \cite{balakrishnan2018synthesizing}, the different body parts of a person are segmented and moved to the target pose by part-specific learnable affine transformations, which are applied at the pixel level. The deformable GAN from \cite{siarohin2018deformable} is a U-net \cite{ronneberger2015u} generator whose skip connections are deformed by part-specific affine transformations. These transformations are computed from the source and target pose information. Instead, \cite{dong2018soft} use the convolutional geometric matcher from \cite{rocco2017convolutional} to learn a thin-plate-spline (TPS) transform between the source human parsing and a synthesized target parsing, and align the deep feature maps of an encoder-decoder generator. 

\paragraph{\textbf{Appearance transfer.}} Close to the virtual try-on task, there is a body of work on human appearance transfer. Given two images of different persons, the goal is to transfer the appearance of a part  of the person A on the person B. Approaches using pose and appearance disentanglement mentioned in the previous section \cite{lorenz2019unsupervised, ma2018disentangled} fit this task but others are specifically designed for it. SwapNet \cite{raj2018swapnet} propose a dual path network to generate a new human parsing of the reference person and region of interest pooling to transfer the texture. In \cite{wu2018m2e}, the method relies on DensePose information \cite{alp2018densepose}, which provides a 3D surface estimation of a human body, to perform a warping and align the two persons. The transfer is then done with segmentation masks and refinement networks.
\paragraph{\textbf{Virtual try-on.}} Most of the approaches for a virtual try-on system come from computer graphics and rely on 3D measurements or representations. Drape \cite{guan2012drape} learns a deformation model to render clothes on 3D bodies of different shapes. In \cite{hahn2014subspace}, Hahn et al. use subspace methods to accelerate physics-based simulations and generate realistic wrinkles.  ClothCap \cite{pons2017clothcap} aligns a 3D cloth-template to each frame of a sequence of 3D scans of a person in motion. However, these methods are targetting the dressing of virtual avatars, e.g for the gaming or movie industry.

The task we are interested in is the one introduced in CAGAN \cite{jetchev2017conditional} and further studied by VITON \cite{han2018viton} and CP-VTON \cite{wang2018toward}, which we defined in the problem statement. In CAGAN \cite{jetchev2017conditional}, Jetchev et al. propose a cycle-GAN approach that requires three images as input: the reference person, the cloth worn by the person and the target in-shop cloth. Thus, it limits its practical uses. VITON \cite{han2018viton} proposes to learn a generative composition between the warped cloth and a coarse result. The warping is done with a non-parametric geometric transform \cite{belongie2002shape}. To improve this model, CP-VTON \cite{wang2018toward} incorporates a learnable geometric matcher $STN$ \cite{rocco2017convolutional} which aligns $c$ with $p$. 
\section{Our approach}
\begin{figure}
\centering
    \includegraphics[width=\linewidth]{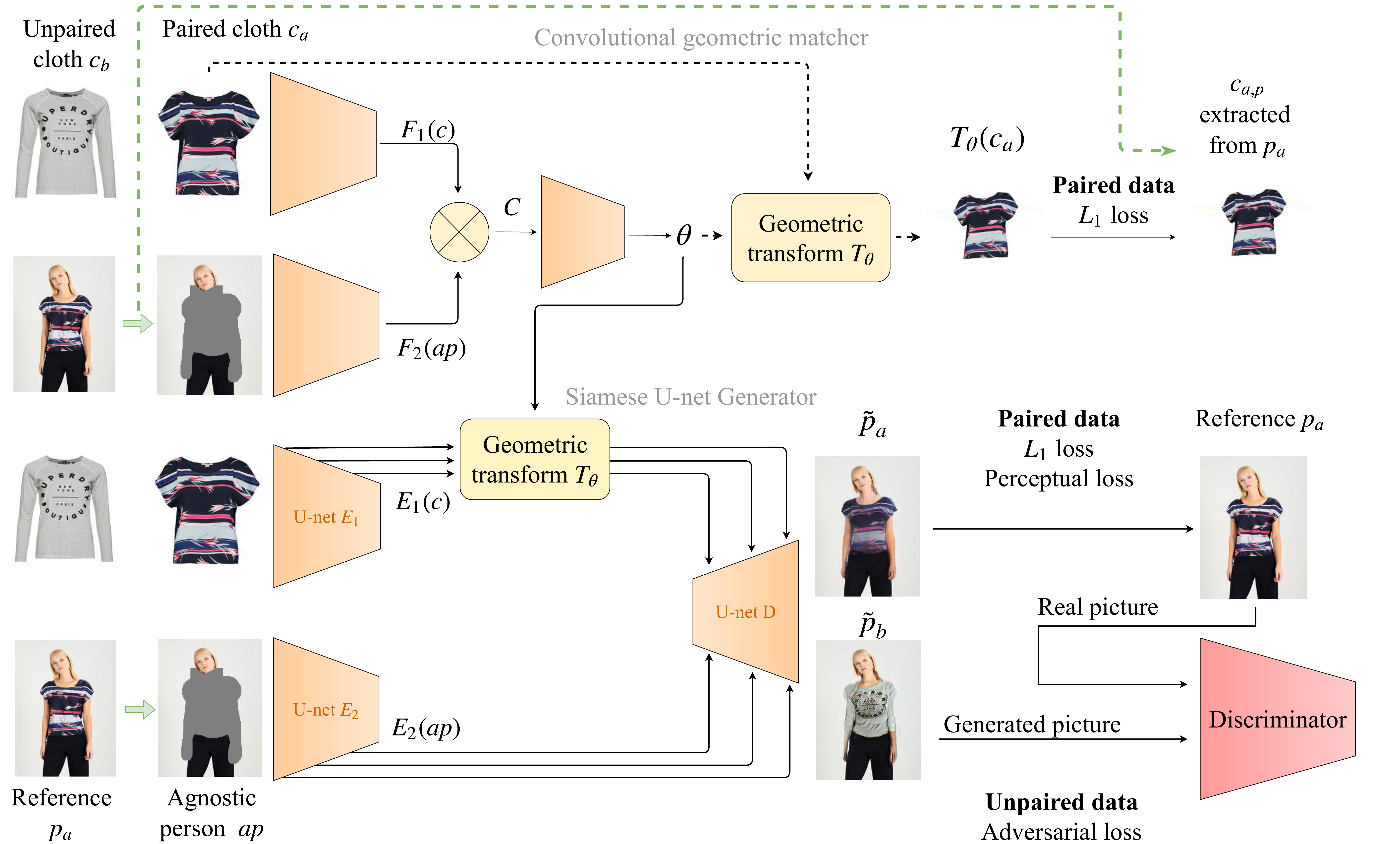}
    \caption{\label{fig:architecture} WUTON : our proposed end-to-end warping U-net architecture. Dotted arrows correspond to the forward pass only performed during training. Green arrows are the human parser. The geometric transforms share the same parameters but do not operate on the same spaces. The different training procedure for paired and unpaired pictures is explained in section \ref{sec:training}.}
\end{figure}
Our task is to build a virtual try-on system that is able to fit a given in-shop cloth on a reference person. We propose a novel architecture trainable end-to-end and composed of two existing modules, \ie{} a convolutional geometric matcher $STN$ \cite{rocco2017convolutional} and a U-net \cite{ronneberger2015u} generator $G$ whose skip connections are deformed by $STN$. The joint training of $G$ and $STN$ allows us to generate realistic deformations that help to synthesize high-quality images. Also, we use an adversarial loss to make the training procedure closer to the actual use of the system which is  to replace clothes in the unpaired situation. In previous works \cite{han2018viton, wang2018toward, dong2019towards}, the generator is only trained to reconstruct images with supervised triplets (\textit{ap}, \textit{c}, \textit{p}) extracted from the \emph{paired}. Thus, when generating images in the test-setting, it can struggle to generalize and to warp clothes different from the one worn by the reference person. The adversarial training allows us to train our network in the test-setting, where one wants to fit a cloth on a reference person wearing another cloth. 

\subsection{Warping U-net}
Our warping U-net is composed of two connected modules, as shown in Fig.\ref{fig:architecture}. The first one is a convolutional geometric matcher, which has a similar architecture as \cite{rocco2017convolutional, wang2018toward}. It outputs the parameters $\theta$ of a geometric transformation, a TPS transform in our case. This geometric transformation aligns the in-shop cloth image with the reference person. However, in contrast to previous work \cite{han2018viton,wang2018toward,dong2019towards}, we use the geometric transformation on the features maps of the generator rather than at a pixel-level. Thus, we learn to deform the feature maps that pass through the skip connections of the second module, a U-net \cite{ronneberger2015u} generator which synthesizes the output image $\Tilde{p}$. 

The architecture of the convolutional geometric matcher is taken from CP-VTON \cite{wang2018toward}, which reuses the generic geometric matcher from \cite{rocco2017convolutional}. It is composed of two feature extractors $F_1$ and $F_2$, which are standard convolutional neural networks. The local vectors of feature maps $F_1(c)$ and $F_2(ap)$ are then L2-normalized and a correlation map $C$ is computed as follows : 
\begin{equation}
     C_{ijk} = F_{1_{i,j}}(c) \cdot F_{2_{m,n}}(c)
\end{equation}
where k is the index for the position (m, n). This correlation map captures dependencies between distant locations of the two feature maps, which is useful to align the two images. $C$ is the input of a regression network, which outputs the parameters $\theta$ and allows to perform the geometric transformation $T_{\theta}$. We use TPS transformations \cite{bookstein1989principal}, which generate smooth sampling grids given control points. %providing deformation constraints. 
Since we transform deep feature maps of a U-net generator, we generate a sampling grid for each scale of the U-net with the same parameters $\theta$. 

The input of the U-net generator is also the tuple of pictures $(ap,c)$. Since these two images are not spatially aligned, we cannot simply concatenate them and feed a standard U-net. To alleviate this, we use two different encoders $E_1$ and $E_2$ processing each image independently and with non-shared parameters. Then, the feature maps of the in-shop cloth $E_1(c)$ are transformed at each scale $i$: $E^i_1(c) = T_{\theta} (E^i_1(c))$. Then, the feature maps of the two encoders are concatenated and feed the decoder. With aligned feature maps, the generator is able to compose them and to produce realistic results. Because we simply concatenate the feature maps and let the U-net decoder compose them instead of enforcing a pixel-level composition, experiments will show that it has more flexibility and can produce more natural results. We use instance normalization in the U-net generator, which is more effective than batch normalization \cite{ioffe2015batch} for image generation \cite{ulyanov2017improved}. 

\subsection{Training procedure}
\label{sec:training}
Along with a new architecture for the virtual try-on task (Fig. \ref{fig:architecture}), we also propose a new training procedure, i.e. a different data representation and an adversarial loss for unpaired images.  

While previous works use a rich person representation with more than 20 channels representing human pose, body shape and the RGB image of the head, we only mask the upper-body of the reference person. Our agnostic person representation $ap$ is thus a 3-channel RGB image with a masked area. We compute the upper-body mask from pose and body parsing information provided by a pre-trained neural network from \cite{liang2019look}. Precisely, we mask the areas corresponding to the arms, the upper-body cloth and a fixed bounding box around the neck keypoint. However, we show in an ablation study that our method is not sensitive to non-accurate masks at inference time since it can generate satisfying images with simple bounding box masks. 

Using the dataset from \cite{dong2019towards}, we have pairs of in-shop cloth image $c_a$ and a person wearing the same cloth $p_a$. Using a human parser and a human pose estimator, we generate $ap$. From the parsing information, we can also isolate the cloth on the image $p_a$ and get $c_{a,p}$, the cloth worn by the reference person. Moreover, we get the image of another in-shop cloth, $c_b$. The inputs of our network are the two tuples $(ap, c_a)$ and $(ap, c_b)$. The outputs are respectively $(\Tilde{p_a}, \theta_a)$ and $(\Tilde{p_b}, \theta_b)$.

The cloth worn by the person $c_{a,p}$ allows us to guide directly the geometric matcher with a $L_1$ loss:
\begin{equation}
    L_{warp} = \lVert T_{\theta_a} (c) - c_{a,p} \rVert_1
\end{equation}
The image $p$ of the reference person provides a supervision for the whole pipeline. Similarly to CP-VTON \cite{wang2018toward}, we use two different losses to guide the generation of the final image $\Tilde{p}_a$, the pixel-level $L_1$ loss $\lVert \Tilde{p}_a - p_a \rVert_1$ and the perceptual loss \cite{johnson2016perceptual}. 
We focus on $L_1$ losses since they are known to generate less blur than $L_2$ for image generation \cite{zhao2016loss}.
The latter consists of using the features extracted with a pre-trained neural network, VGG \cite{simonyan2014very} in our case. Specifically, our perceptual loss is: 
\begin{equation}
    L_{perceptual} = \sum_{i = 1}^{5} \lVert \phi_i(\Tilde{p}_a) - \phi_i(p_a) \rVert_1
\end{equation}
where $\phi_i (I)$ are the feature maps of an image I extracted at the i-th layer of the VGG network. 
Furthermore, we exploit adversarial training to train the network to fit $c_b$ on the same agnostic person representation $ap$, which is extracted from a person wearing $c_a$. This is only feasible with an adversarial loss, since there is no available ground-truth for this pair $(ap, c_b)$. Thus, we feed the discriminator with the synthesized image $\Tilde{p}_b$ and real images of persons from the dataset. This adversarial loss is also back-propagated to the convolutional geometric matcher, which allows to generate much more realistic spatial transformations. We use the relativistic adversarial loss \cite{jolicoeur-martineau2018} with gradient-penalty \cite{gulrajani2017improved,arjovsky2017wasserstein}, which trains the discriminator to predict relative realness of real images compared to synthesized ones.  
Finally, the objective function of our network is : 
\begin{equation}
     L = \lambda_{w} L_{warp} + \lambda_{p} L_{perceptual} + \lambda_{L_1} L_1 + \lambda_{adv} L_{adv}
\end{equation}
We use the Adam optimizer \cite{kingma2014adam} to train our network. 

\section{Experiments and analysis}

\begin{figure}[ht]
\centering
\begin{tabular}{cccc}
Reference & Target & CP-VTON & WUTON \\
person & cloth &  & \\
\subfigure{\includegraphics[width=0.20\linewidth]{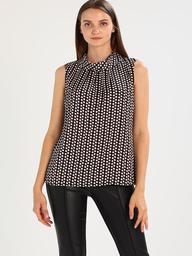}} & 
\subfigure{\includegraphics[width=0.20\linewidth]{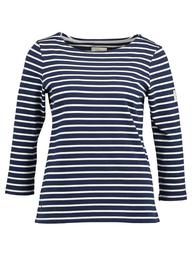}} & 
\subfigure{\includegraphics[width=0.20\linewidth]{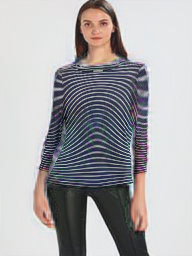}} & 
\subfigure{\includegraphics[width=0.20\linewidth]{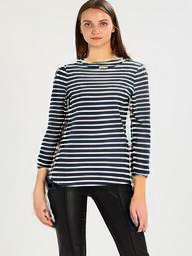}} \\
\subfigure{\includegraphics[width=0.20\linewidth]{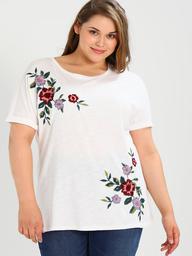}} & 
\subfigure{\includegraphics[width=0.20\linewidth]{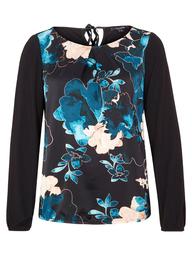}} & 
\subfigure{\includegraphics[width=0.20\linewidth]{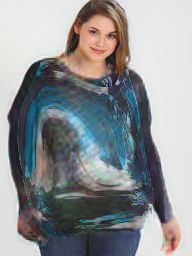}} & 
\subfigure{\includegraphics[width=0.20\linewidth]{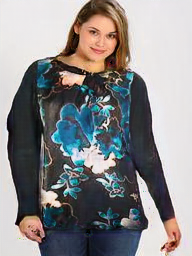}} \\
\subfigure{\includegraphics[width=0.20\linewidth]{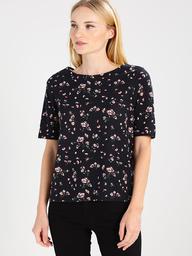}} & 
\subfigure{\includegraphics[width=0.20\linewidth]{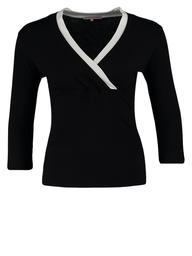}} & 
\subfigure{\includegraphics[width=0.20\linewidth]{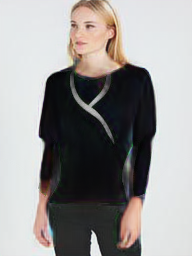}} & 
\subfigure{\includegraphics[width=0.20\linewidth]{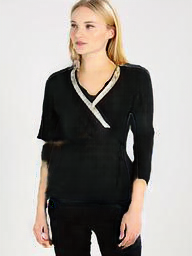}} \\
\end{tabular}
\label{fig:comparison}
\caption{Comparison of our method with CP-VTON \cite{wang2018toward}, the state-of-the-art for the virtual try-on task. More examples and higher resolution pictures are provided in Appendix.}
\end{figure}

\begin{figure}[ht]
\centering\begin{tabular}{c@{}c@{ }c@{ }c@{ }c@{}c@{}c@{}c@{}}
Reference & Target & Unpaired & Paired & No adv. & Not\\
 person & cloth & adv. loss & adv. loss & loss & end-to-end \\
%\hspace*{-3em}
\subfigure{\includegraphics[width=0.12\linewidth]{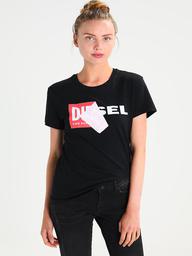}} & 
\subfigure{\includegraphics[width=0.12\linewidth]{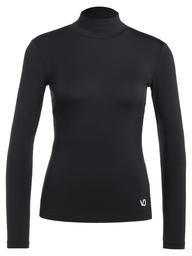}} & 
\subfigure{\includegraphics[width=0.12\linewidth]{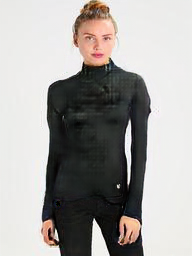}} & 
\subfigure{\includegraphics[width=0.12\linewidth]{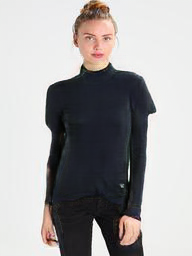}} &
\subfigure{\includegraphics[width=0.12\linewidth]{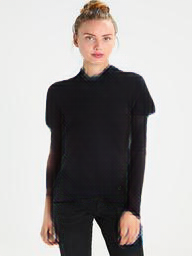}} &
\subfigure{\includegraphics[width=0.12\linewidth]{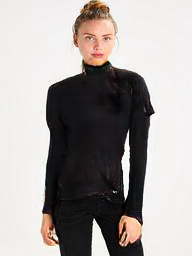}} \\[-0.8ex]
\subfigure{\includegraphics[width=0.12\linewidth]{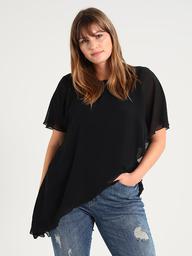}} & 
\subfigure{\includegraphics[width=0.12\linewidth]{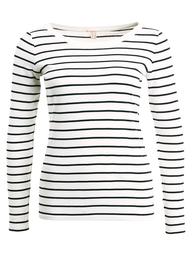}} & 
\subfigure{\includegraphics[width=0.12\linewidth]{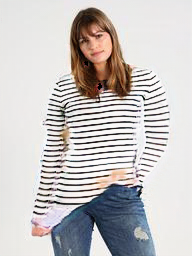}} & 
\subfigure{\includegraphics[width=0.12\linewidth]{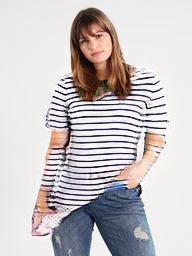}} &
\subfigure{\includegraphics[width=0.12\linewidth]{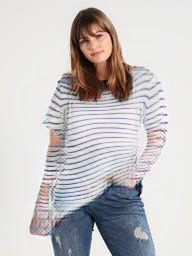}} &
\subfigure{\includegraphics[width=0.12\linewidth]{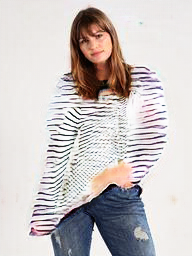}} \\[-0.8ex]
\subfigure{\includegraphics[width=0.12\linewidth]{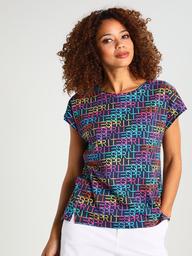}} & 
\subfigure{\includegraphics[width=0.12\linewidth]{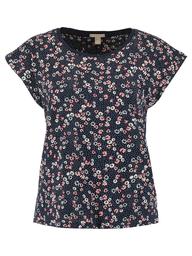}} & 
\subfigure{\includegraphics[width=0.12\linewidth]{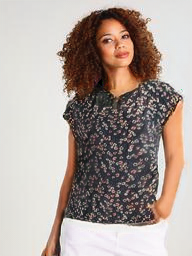}} & 
\subfigure{\includegraphics[width=0.12\linewidth]{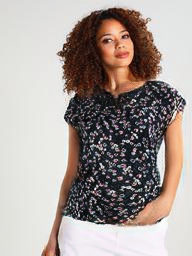}} &
\subfigure{\includegraphics[width=0.12\linewidth]{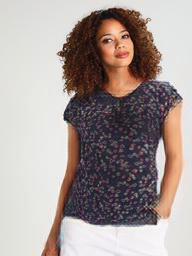}} &
\subfigure{\includegraphics[width=0.12\linewidth]{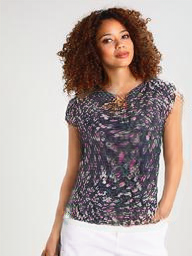}} \\
\end{tabular}
\caption{Our unpaired adversarial loss function improves the performance of our generator in the case of significant shape changes from the source cloth to the target cloth. Specifically, when going from short sleeves to long sleeves, it tends to gum the shape of the short sleeves. With the paired adversarial loss, we do not observe this phenomenon since the case never happens during training.}
\label{fig:loss}
\end{figure}

\begin{figure}[ht]
\centering\begin{tabular}{ccc|ccc}
Masked & Target   & WUTON  & Reference  & Target   & WUTON  \\
 image & cloth   &  & person & cloth &  \\
\subfigure{\includegraphics[width=0.11\linewidth]{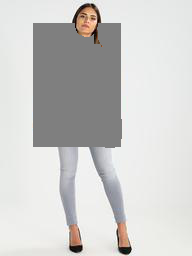}} & 
\subfigure{\includegraphics[width=0.11\linewidth]{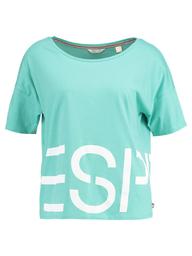}} & 
\subfigure{\includegraphics[width=0.11\linewidth]{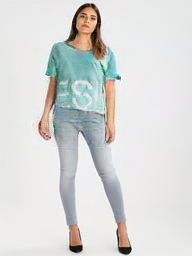}} &
\subfigure{\includegraphics[width=0.11\linewidth]{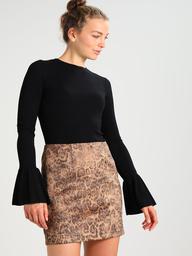}} & 
\subfigure{\includegraphics[width=0.11\linewidth]{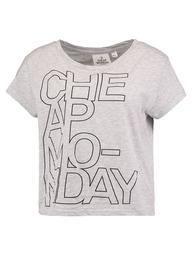}} & 
\subfigure{\includegraphics[width=0.11\linewidth]{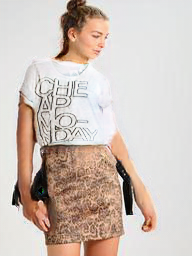}} \\
\subfigure{\includegraphics[width=0.11\linewidth]{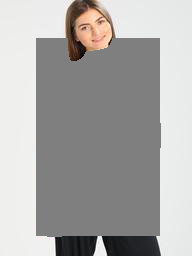}} & 
\subfigure{\includegraphics[width=0.11\linewidth]{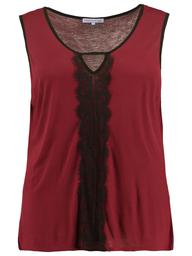}} & 
\subfigure{\includegraphics[width=0.11\linewidth]{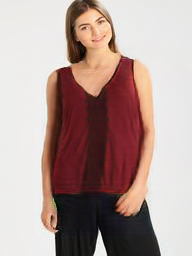}} &
\subfigure{\includegraphics[width=0.11\linewidth]{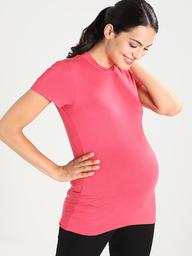}} & 
\subfigure{\includegraphics[width=0.11\linewidth]{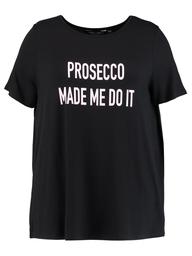}} & 
\subfigure{\includegraphics[width=0.11\linewidth]{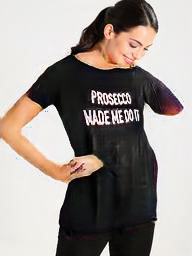}} \\
\end{tabular}
\label{fig:failure_cases}
\caption{Left:  Our method can handle low-quality masks at cost of generic arm pose. 
Right: Some common failure cases of our method. Detection of initial cloth can fail beyond the capacity of our U-net generator (first row), and uncommon poses are not properly rendered (second row).}
\end{figure}

We first describe the dataset. Then, we compare our approach with CP-VTON \footnote{We use the public implementation from \href{https://github.com/sergeywong/cp-vton}{https://github.com/sergeywong/cp-vton}.} \cite{wang2018toward}, the current state-of-the-art for the virtual try-on task. We present visual and quantitative results proving that WUTON significantly outperforms the current state-of-the-art for a virtual try-on.  Finally, we describe the impact of each main component of our approach in an ablation study and show that WUTON can also generate high-quality images with a non-accurate mask at inference time. 
\subsection{Dataset}
We use the \textit{Image-based Multi-pose Virtual try-on} dataset\footnote{The dataset is available at: \href{http://sysu-hcp.net/lip/overview.php}{http://sysu-hcp.net/lip/overview.php}} from MG-VTON \cite{dong2019towards}. This dataset contains 35,687/13,524 person/cloth images at (256,192) resolution. For each in-shop cloth image, there are multiple images of a model wearing the given cloth from different views and in different poses. We remove images tagged as back images since the in-shop cloth image is only from the front. We process the images with a neural human parser and pose estimator, specifically the joint body parsing and pose estimation (JPP) network\footnote{This human parser and pose estimator is open-source and available at: \href{https://github.com/Engineering-Course/LIP_JPPNet}{https://github.com/Engineering-Course/LIP\_JPPNet}.} \cite{liang2019look,gong2017look}.

\subsection{Visual results}
Visual results of our method and CP-VTON are shown in Fig. \ref{fig:comparison}. CP-VTON has trouble to realistically deform and render complex patterns like stripes or flowers. Control points of the $T_\theta$ transform are visible and lead to unrealistic curves and deformations on the clothes. Also, the edges of cloth patterns and body contours are blurred. 

Our method surpasses the previous state-of-the-art on different challenges. On the two first rows, our method generates spatial transformations of a much higher visual quality, which is specifically visible for stripes (2nd row). It is able to preserve complex visual patterns of clothes and presents less blur than CP-VTON on the edges. Also, it can distinguish the relevant parts of the in-shop cloth image (3rd row). Generally, our method generates results of high visual quality while preserving the characteristics of the target cloth. We also show some failure cases in Fig. \ref{fig:failure_cases}. Problems happen when the human parser does not properly detect the original cloth or when models have uncommon poses. 

\subsection{LPIPS metric}
To further evaluate our method, we use the linear perceptual image patch similarity (LPIPS) metric developed in \cite{zhang2018unreasonable}. This metric is very similar to the perceptual loss we use in training (see Section 3.2) since the idea is to use the feature maps extracted by a pre-trained neural network to quantify the perceptual difference between two images. Different from the basic perceptual loss, they first unit-normalize each layer in the channel dimension and then rescale by learned weights $w_i \in \mathbb{R}^{C_i}$ :
\begin{equation}
    LPIPS(\Tilde{p}_a,p_a) = \sum_{i = 1}^{5} \frac{1}{H_i W_i} \sum_{h,w} \lVert w_i \cdot ( \phi^i_{h,w}(\Tilde{p}_a) - \phi^i_{h,w}(p_a) ) \rVert_2^2
\end{equation}
where $\phi^i_{h,w}$ is the unit-normalized vector at i-th layer and (h,w) is the spatial location extracted by a neural network, AlexNet \cite{krizhevsky2014one} in their case. 

We evaluate the LPIPS on the test set. We can only use this method in the paired setting since there is no available ground-truth in the unpaired setting. Thus, it does not exactly evaluate the real task we aim for. Results are shown in Table \ref{table:lpips}. Our approach significantly outperforms the state-of-the-art on this metric.
Here the best model uses adversarial loss on paired data, but visual investigation suggests that the unpaired adversarial loss is better in the real use case of our work. 
We evaluate CP-VTON \cite{wang2018toward} on their agnostic person representation $ap_{viton}$ (20 channels with RGB image of head and shape/pose information) and on our $ap_{wuton}$. 

\begin{table}[tb]
    \caption{LPIPS metric on paired setting. Lower is better, $\pm$ reports std. dev.}
     \label{table:lpips}
\begin{multicols}{2}
   %\begin{minipage}[t]{.5\linewidth}
    %  \centering
        \begin{tabular}{lll}
            \toprule
            Method    & LPIPS  \\
            \midrule
            CP-VTON on $ap_{viton}$  & 0.182 $\pm$ 0.049    \\
            CP-VTON on $ap_{wuton}$ & 0.131   $\pm$ 0.058 \\
            WUTON & 0.101 $\pm$ 0.047 \\
            \midrule 
            \textbf{Impact of loss} \\
            \textbf{functions on WUTON:} \\
            W/o adv. loss    & 0.107 $\pm$ 0.049  \\
            W. paired adv. loss &  0.099  $\pm$ 0.046\\
            Not end-to-end & 0.112 $\pm$ 0.053\\
            \bottomrule
          \end{tabular}
   % \end{minipage}%
   \newpage
    %\begin{minipage}[t]{.5\linewidth}
    %  \centering
        \begin{tabular}{lll}
            \toprule
            Method    & LPIPS \\
            \midrule
            \textbf{Impact of composition} \\
            \textbf{on WUTON:} \\
            W. composition & 0.105 $\pm$ 0.047\\
            \midrule 
            \textbf{Impact of mask quality} \\
            \textbf{box masked person:} \\
            CP-VTON & 0.185 $\pm$ 0.078 \\
            WUTON & 0.151 $\pm$ 0.069 \\
            \bottomrule
        \end{tabular}
%    \end{minipage}
\end{multicols}
\end{table}

\subsection{Ablation studies}
To prove the effectiveness of our approach, we perform several ablation studies.
In Fig. \ref{fig:loss}, we show visual comparisons of different variants of our approach: our WUTON with unpaired adversarial loss; with an adversarial loss on paired data (i.e the adversarial loss is computed with the same synthesized image as the L1 and VGG losses); without the adversarial loss; without back-propagating the loss of the synthesized images ($L_1, L_{perceptual}, L_{adv}$) to the geometric matcher. 

The adversarial loss  generates sharper images and improves the contrast. This is confirmed by the LPIPS metric in Table \ref{table:lpips} and with visual results in Fig. \ref{fig:loss}. With the unpaired adversarial setting, the system better handles large variations between the shape of the cloth worn by the person and the shape of the new cloth. 
The results in Fig. \ref{fig:loss} as well as the LPIPS score in Table \ref{table:lpips} show the importance of our end-to-end learning of geometric deformations. When the geometric matcher only benefits from $L_{warp}$, it only learns to align $c$ with the masked area in $ap$. However, it does not preserve the inner structure of the cloth. Back-propagating the loss computed on the synthesized images $\Tilde{p}$ alleviates this issue. 
Finally, our approach removes the need for learning a composition between the warped cloth and a coarse result. To prove it, we re-design our U-net to generate a coarse result and a composition mask. The synthesized image is then the composition between the coarse result and the warped cloth. With this configuration, the LPIPS score slightly decreases.   

We also show that our method can  generate realistic results if the human parser is not accurate at inference time. Hence, we train and test our method with the upper-body of the person masked by a gray bounding box. It is to be noted that we still require the accurate human parsing during training for the warping loss $L_{warp}$. 

We also tried to learn the architecture without using $L_{warp}$, but this lead to major trouble with the convergence of the networks. 
On one hand, a possible future direction is to try to reduce the dependency to the human parser by learning the segmentation in a self-supervised way (providing several pictures of the same cloth on different models or using videos). On the other hand, the presence of this parser can ease the handling of pictures with complex backgrounds.

\section{Conclusion}
In this work, we propose an architecture trainable end-to-end which combines a U-net with a convolutional geometric matcher and significantly outperforms the state-of-the-art for the virtual try-on task. The end-to-end training procedure with an unpaired adversarial loss allows to generate realistic geometric deformations, synthesize sharp images with proper contrast and preserve complex patterns, including stripes and logos. 
\bibliographystyle{unsrt}
\bibliography{references}

\cleardoublepage

\appendix

\section{Appendix}
\subsection{Implementation details}
\textbf{Convolutional geometric matcher.} To extract the feature maps, we apply five times one standard convolution layer followed by a 2-strided convolution layer which downsamples the maps. The depth of the feature maps at each scale is (16,32,64,128,256). The correlation map is then computed and feeds a regression network composed of two 2-strided convolution layers, two standard convolution layers and one final fully connected layer predicting a vector $\theta \in \mathbb{R}^{50}$. We use batch normalization \cite{ioffe2015batch} and relu activation. The parameters of the two feature maps extractors are not shared. 

\textbf{Siamese U-net generator.} We use the same encoder architecture as in the convolutional geometric matcher, but we store the feature maps at each scale. The decoder has an architecture symmetric to the encoder. There are five times one standard convolution layer followed by an 2-strided deconvolution layer which upsamples feature maps. After a deconvolution, the feature maps are concatenated with the feature maps passed through skip connections. In the generator, we use instance normalization, which shows better results for image and texture generation \cite{ulyanov2017improved}, with relu activation. 

\textbf{Discriminator.} We adopt the fully convolutional discriminator from Pix2Pix \cite{isola2017image}, but with five downsampling layers instead of three in the original version. Each of it is a 2-strided convolution layer with batch normalization and a leaky relu activation. 

\textbf{Adversarial loss.} We use the relativistic formulation of the adversarial loss \cite{jolicoeur-martineau2018}. In this formulation, the discriminator is trained to predict that real images are more real than synthesized ones, rather than trained to predict that real images are real and synthesized images are synthesized. 

\textbf{Optimization.} We use the Adam optimizer \cite{kingma2014adam} with $\beta_1 = 0.5$, $\beta_2 = 0.999$, a learning rate of $10e^{-3}$ and a batch size of 8. Also, we use $\lambda_p = \lambda_{L1} = \lambda_w = \lambda_{adv} = 1$. 

\textbf{Hardware.} We use a NVIDIA Tesla V100 with 16GB of RAM. The training takes around 3 days. For inference, WUTON processes a mini-batch of 4 images in $\sim$0.31s.

\end{document}